\documentclass[letterpaper, 10 pt, conference]{ieeeconf}  

\IEEEoverridecommandlockouts                              

\usepackage{cite}
\usepackage{graphics} 
\usepackage{epsfig} 
\usepackage{mathptmx} 
\usepackage{times} 
\usepackage{amsmath} 
\usepackage{amssymb}  

\usepackage{multirow}
\usepackage{pifont}

\newcommand{\omni}{{\sc Omniverse}}

\usepackage{xcolor}

\usepackage{algpseudocode}
\usepackage{algorithm}

\usepackage[symbol]{footmisc}

\title{\LARGE \bf OpenD: A Benchmark for Language-Driven Door and Drawer Opening}

\author{Yizhou Zhao$^{1}$, Qiaozi Gao$^{2}$, Liang Qiu$^2$, Govind Thattai$^{2}$, Gaurav S. Sukhatme$^{2,3}$
\thanks{$^{1}$Y. Zhao is with the Department of Statistics, University of California, Los Angeles
        {\tt\small yizhouzhao@g.ucla.edu}}%
\thanks{$^{2}$Q. Gao, L. Qiu, G. Thattai, and G. Sukhatme are with Amazon Alexa AI {\tt\small \{qzgao, liangqxx, thattg, sukhatme\}@amazon.com}}%
\thanks{$^{3}$G. Sukhatme is with the Computer Science Department, University of Southern California
        {\tt\small gaurav@usc.edu}}%
}

\begin{document}

\maketitle
\thispagestyle{empty}
\pagestyle{empty}

\begin{abstract}

We introduce \textsc{OpenD}, a benchmark for learning how to use a hand to open cabinet doors or drawers in a photo-realistic and physics-reliable simulation environment driven by language instruction. To solve the task, we propose a multi-step planner composed of a deep neural network and rule-base controllers. The network is utilized to capture spatial relationships from images and understand semantic meaning from language instructions. Controllers efficiently execute the plan based on the spatial and semantic understanding.
We evaluate our system by measuring its zero-shot performance in test data set. Experimental results demonstrate the effectiveness of decision planning by our multi-step planner for different hands, while suggesting that there is significant room for developing better models to address the challenge brought by language understanding, spatial reasoning, and long-term manipulation. We will release \textsc{OpenD} and host challenges to promote future research in this area.\footnote{This work has been submitted to the IEEE for possible publication. Copyright may be transferred without notice, after which this version may no longer be accessible.}

\end{abstract}

\section{Introduction}

With the development of simulation engines, Embodied AI research~\cite{duan2022survey} is contributing to progress in intelligent robotic systems. Thanks to manipulation benchmarks~\cite{zheng2022vlmbench,mu2021maniskill,shridhar2020alfred,ehsani2021manipulathor,robomimic2021}, new models and algorithms have emerged to help robotics research in object manipulation and overcome the domain gap between virtual and  physical spaces.

While most existing benchmarks in simulation attempt to cover a wide range of tasks, the tasks themselves are simplified. The simplification is usually achieved by changing object sizes~\cite{mu2021maniskill,zheng2022vlmbench}, de-emphasizing collisions~\cite{shridhar2020alfred, srivastava2022behavior}, or abstracting away the details of grasping~\cite{shridhar2020alfred, ehsani2021manipulathor}. By assuming binary or discrete object and robot states~\cite{szot2021habitat, srivastava2022behavior, ehsani2021manipulathor} the system is not forced to reason about object geometry and physical laws. Performing training in a simple, background-free environment~\cite{kumar2015mujoco, lin2020softgym} also limits any learned model's ability to expand to real applications.

To remedy some of the limitations of existing benchmarks, we present \textsc{OpenD}, a benchmark for learning hand manipulation skills to mobilize articulated objects based on visual and language input. We focus on one task: opening a cabinet (from its door or drawer), but we bring significant realism and scalability, both in terms of visualization and physics.

\begin{figure}
    \centering
    \includegraphics[width = 0.5\textwidth]{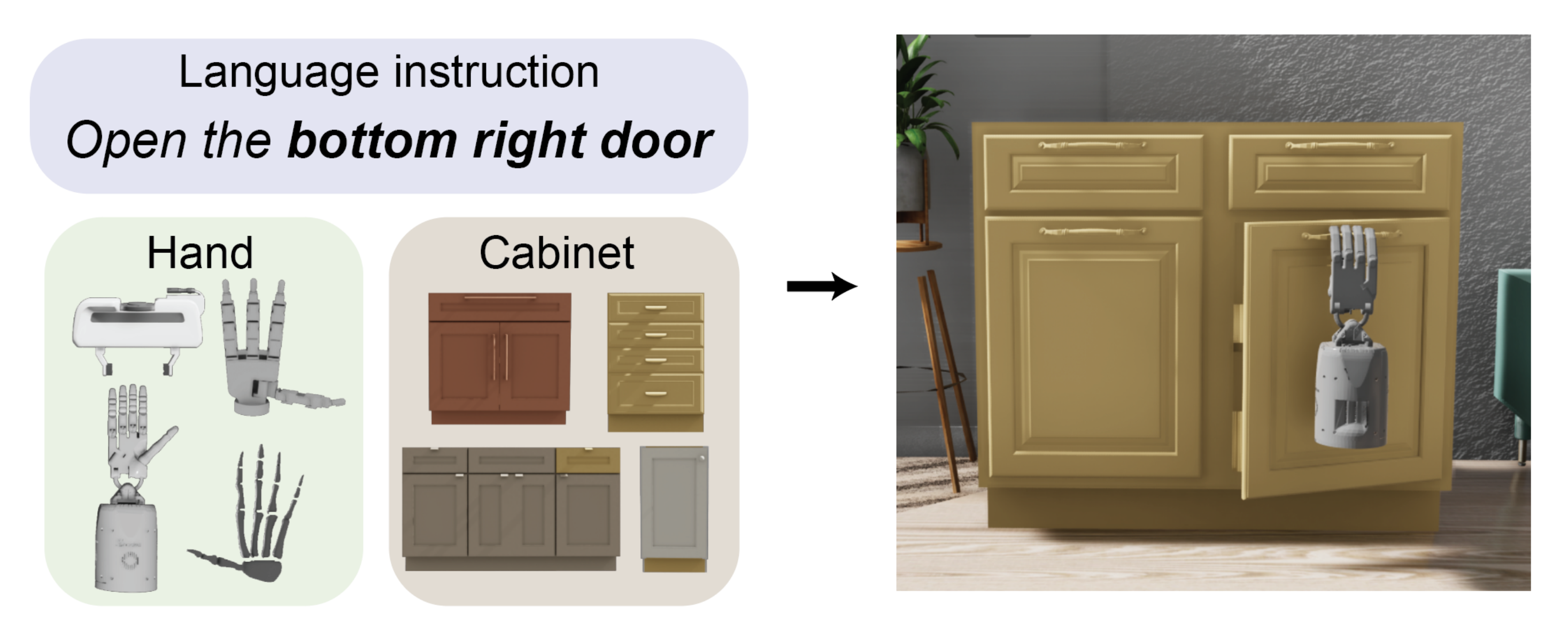}
    \caption{{\bf Problem setting.} In a simulated scene, the task is to open a cabinet door or drawer by hand corresponding to the language instruction and the camera image. \textsc{OpenD} provides 174 different cabinets, 372 pieces of language instructions, and 4 types of hands. }
    \label{fig:first_look}
\end{figure}

\textsc{OpenD} imports high-topology objects with varying geometry and high-quality room backgrounds, as shown in Figure~\ref{fig:first_look}. It currently includes hundreds of automatically processed cabinets and several manually designed rooms with randomized floor and wall materials.
To simulate physics, \textsc{OpenD} considers both collisions and friction between hand fingers and the cabinet without simplifying their geometries. We provide four typical robot hands for the manipulation task, posing the challenge of controlling the hand via all its joints. Besides, to encourage language-instructed learning, hundreds of instructions are generated automatically from parsing the cabinet's geometry. Therefore, a concurrent challenge is to correctly understand and interpret language instructions.

We demand an efficient and stable model to overcome difficulties brought by long-horizon planning and high-dimensional control. To address these challenges, we propose a baseline model using a multi-step planner, which combines network training and motion planning. The network leverages the advantages of the Faster R-CNN~\cite{ren2015faster} and CLIPort~\cite{shridhar2022cliport} to perform recognition tasks. Combined with a set of pre-defined controllers, our baseline model enables precise hand joint control. 

We evaluate our approach on an unseen test data set. The model only has one chance to execute, and the cabinet is not revealed during training or validation. Only the image of the cabinet obtained from a camera, and the language instruction, are provided.

Experiments show that the models trained for different hands produce varied success rates on test data ranging from $14.6\%$ to $30.3\%$. The results suggests that our baseline can help perform long-term manipulation to control the hand and fingers to open the target cabinet. However, these success rates are far from ideal for this challenge. There exists significant room for developing innovative models to perform long-term and accurate manipulation based on vision and language understanding in this setting.


\section{Related work}
\begin{table*}[t]
\label{tab:comparison}
\begin{center}
\caption{Comparison with other manipulation benchmarks.}
\begin{tabular}{l|ccccccc}
\hline
\textbf{Benchmark} & \textbf{\begin{tabular}[c]{c}Realistic \\grasping\end{tabular}} & \textbf{\begin{tabular}[c]{c}Various\\ cabinet type\end{tabular}} & \textbf{\begin{tabular}[c]{c}Realistic \\ cabinet size\end{tabular}} & \textbf{\begin{tabular}[c]{@{}c@{}}6-DOF\\ hand\end{tabular}} & \textbf{\begin{tabular}[c]{c}Various \\ robot/hand type\end{tabular}} & \textbf{\begin{tabular}[c]{@{}c@{}}Photo-realistic\\ background\end{tabular}} & \textbf{Multi-task}\\ \hline
ManipulaTHOR~\cite{ehsani2021manipulathor}       & \ding{55}                                                                  & \ding{51}                                                                        & \ding{51}                                                                           & \ding{51}                                                               & \ding{55}                                                                            & \ding{51}       & \ding{51}       \\
VLMBench~\cite{zheng2022vlmbench}           & \ding{51}                                                                  & \ding{55}                                                                        & \ding{55}                                                                           & \ding{51}                                                                & \ding{55}                                                                            & \ding{55}         &\ding{51}                                                                         \\
ManiSkill~\cite{mu2021maniskill}          & \ding{51}                                                                  & \ding{51}                                                                        & \ding{55}                                                                           & \ding{51}                                                                & \ding{55}                                                                            & \ding{55}           &\ding{51}                                                                       \\
Calvin~\cite{mees2022calvin}            & \ding{51}                                                                  & \ding{55}                                                                        & \ding{51}                                                                           & \ding{51}                                                                & \ding{55}                                                                            & \ding{55}          &\ding{51}                                                                        \\
Robomimic~\cite{robomimic2021}          & \ding{51}                                                                  & \ding{55}                                                                        & \ding{51}                                                                           & \ding{51}                                                                & \ding{55}                                                                            & \ding{51}             &\ding{51}                                                                     \\ \hline
Ours               & \ding{51}                                                                  & \ding{51}                                                                        & \ding{51}                                                                           & \ding{51}                                                                & \ding{51}                                                                            & \ding{51}                  &\ding{55}                                                                \\ \hline
\end{tabular}
\end{center}

\end{table*}


\textbf{Simulation Environment.} There is a large body of work on simulating indoor household activities for training and evaluating AI agents~\cite{kolve2017ai2, puig2018virtualhome,li2021igibson, szot2021habitat}. Most of these simulators emulate high-level instructions and post effects of agent behaviors using simplified state and action representations. Some use simplified abstract discrete action space~\cite{shridhar2020alfred} which can reduce the task's difficulty. However, models trained in the setting without any awareness of the low-level geometry and dynamics of the objects are limited in their ability to transfer to  real-world application. For example, grasping is often simplified by attaching a nearby item to the gripper~\cite{ehsani2021manipulathor, shridhar2020alfred, srivastava2022behavior, szot2021habitat}. By contrast, in this work hands are controlled at the joint level  and friction-based grasping powered by a state-of-the-art physics engine PhysX5.0~\cite{physx5}. 

\textbf{Manipulation Task}. Mastering a manipulation skill for a robot usually requires an understanding of vision, language, and robotics. Recently, this field has attracted much attention across disciplines. Beyond applying imitation learning or reinforcement learning to train the robot to grasp and manipulate objects~\cite{mees2022calvin, mu2021maniskill}, recent work proposes end-to-end networks that can learn skillful controls that require precise spatial reasoning or language understanding~\cite{shridhar2022cliport}.
With the help of an environment that offers high-performance physics simulation, training an agent for manipulation tasks can be achieved efficiently. 

However, learning a model from image and language input in the simulation environment with continuous states is still considered challenging~\cite{duan2022survey, mees2022matters}, as the simulation engine needs to provide rendering and simulation results constantly. Therefore, for manipulation benchmarks, compromise often occurs by giving the model full knowledge of the environment and objects for easier training~\cite{mees2022matters}, reducing the difficulty of grabbing items~\cite{shridhar2020alfred}, and providing a limited number of items without varying the material and background for simpler evaluation~\cite{zheng2022vlmbench, mees2022calvin}.    

Our work simulates continuous states for articulated objects and uses the camera sensing and language instructions as model inputs. Without applying abstract grasping or changing the object size, we train our hands  to open the cabinet by hybridizing the end-to-end network and the motion planner. The advantage of the neural network lies in its powerful spatial reasoning capability, and the motion planner helps stable performance based on spatial reasoning.


\textbf{Language Guided Manipulation}. Relating human language to robot actions has been of interest in recent research~\cite{mees2022calvin, lynch2020language, shridhar2020alfred,gao2022dialfred,suglia2021embodied}. Natural language presents specification, providing an intuitive way to refer to abstract concepts concerning spatial, temporal, and causal relationships. 


We focus on language that describes the object type and spatial relationships in our task. Because of multiple interactive parts on the cabinet, language is crucial to address ambiguity. Table \ref{tab:comparison} compares the settings between our \textsc{OpenD} and other benchmarks.


\section{Simulation environment}

This section introduces how we set up the challenge that retains different simulated hands, hundreds of drawers and cabinets, and various scenes with randomized material and backgrounds.

\textbf{Engine.}
We choose \omni~\cite{omniverse} as the platform to design the \textsc{OpenD} challenge. Rigid bodies, soft bodies, articulated objects, and fluids can be efficiently and reliably simulated in \omni. For its Python scripting environment, it is easy to bring open-source and third-party Python libraries. Besides, its ray tracing technology enables powerful rendering effects.

\textbf{Assets.}
The cabinets assets are from SAPIEN PartNet-Mobility dataset~\cite{xiang2020sapien}. As a collection of rigid bodies, robots, and articulated objects, the SAPIEN dataset provides $346$ pieces of storage furniture as cabinets for our challenge. The furniture is well-equipped with detailed rendering material. To build photo-realistic simulation backgrounds, we manually design nine rooms as synthetic indoor scenes with randomized lights, floor materials, wall types, and decorations.

\textbf{Hand.}
A robotic hand, usually programmable, functions similarly to a human hand. \textsc{OpenD} provides control for four representative robot hands in the simulation environment. The first three, \textit{Franka gripper}, \textit{Allegro hand}, and \textit{Shadow hand}, are commercially available robot hands. The last one, \textit{Skeleton hand}, is modeled based on the biological structure of the human hand.

\begin{figure}[h]
    \centering
    \includegraphics[width = 0.5\textwidth]{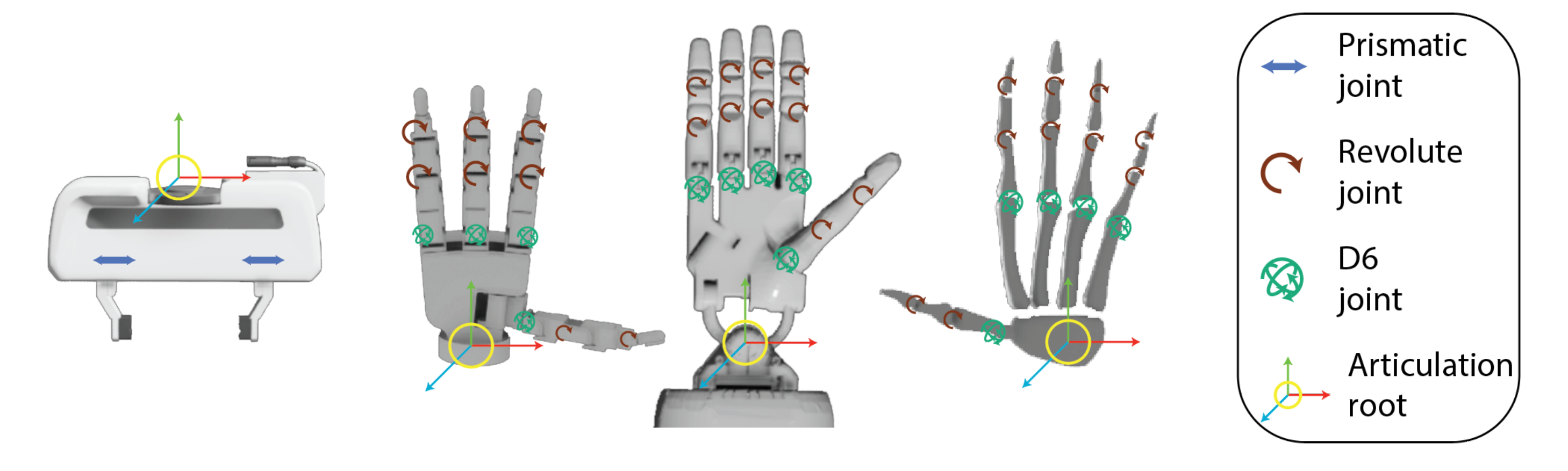}
    \caption{{\bf Hand riggings.} From left to right: Franka gripper, Allegro hand, Shadow hand, and Skeleton hand.}
    \label{fig:hand}
\end{figure}

Figure \ref{fig:hand} sketches how hands are modeled and rigged in \omni. There are three types of joints for hand rigging. The prismatic joint allows two bodies to slide along a common axis; the revolute joint allows two bodies to rotate along a common axis; the D6 joint for the hands enables body parts to spin on the $y$-axis and $z$-axis while locking the rotation on the $x$-axis. 

Table~\ref{tab:hand_dof} lists each hand's joint components and degrees of freedom to control the fingers. Additionally, we need six degrees of freedom to control the position and rotation of the hand root (articulation root). We describe the hand state by its position $p$, rotation $r$, and joint positions $d = \{d_i\}$.  

\begin{table}[h]
\caption{Hand joint components and degrees of freedom.}
\begin{center}
\begin{tabular}{c|ccc|c}
\hline
               & \textbf{\begin{tabular}[c]{@{}c@{}}Prismatic \\ joint\end{tabular}} & \textbf{\begin{tabular}[c]{@{}c@{}}Revolute \\ joint\end{tabular}} & \textbf{\begin{tabular}[c]{@{}c@{}}D6 \\ joint\end{tabular}} & \textbf{DoF} \\ \hline
Franka gripper & 2                                                                   & 0                                                                  & 0                                                            & 2            \\
Allegro hand   & 0                                                                   & 8                                                                  & 4                                                            & 16           \\
Shadow hand    & 0                                                                   & 10                                                                 & 5                                                            & 20           \\
Skeleton hand  & 0                                                                   & 10                                                                 & 5                                                            & 20           \\ \hline
\end{tabular}
\end{center}
\label{tab:hand_dof}
\end{table}

\section{The task}

The challenge of \textsc{OpenD} is to pull open a (cabinet) drawer by $\delta$ meters or rotate a (cabinet) door by $\theta$ degrees, providing that there is a camera placed in front of the cabinet, and a language instruction describing the opening target.

Therefore, the task is to plan the movement of the hand over time, given an image $I$ and language instruction $S$: 
$$
(I, S) \to (p_t, r_t, d_t)
$$

\textbf{Camera sensing.} 
A camera is placed in front of the cabinet with an offset to ensure that the whole cabinet is included in the frame. The camera thus obtains an RGB image. Besides, the camera provides a depth map for the cabinet, specifying the distance between the cabinet and the camera.







\textbf{Language instruction.}
Since a cabinet may have multiple doors and drawers, the language instruction specifies what and where to open.  

\begin{figure}[h] 
    \centering
    \includegraphics[width = 0.5\textwidth]{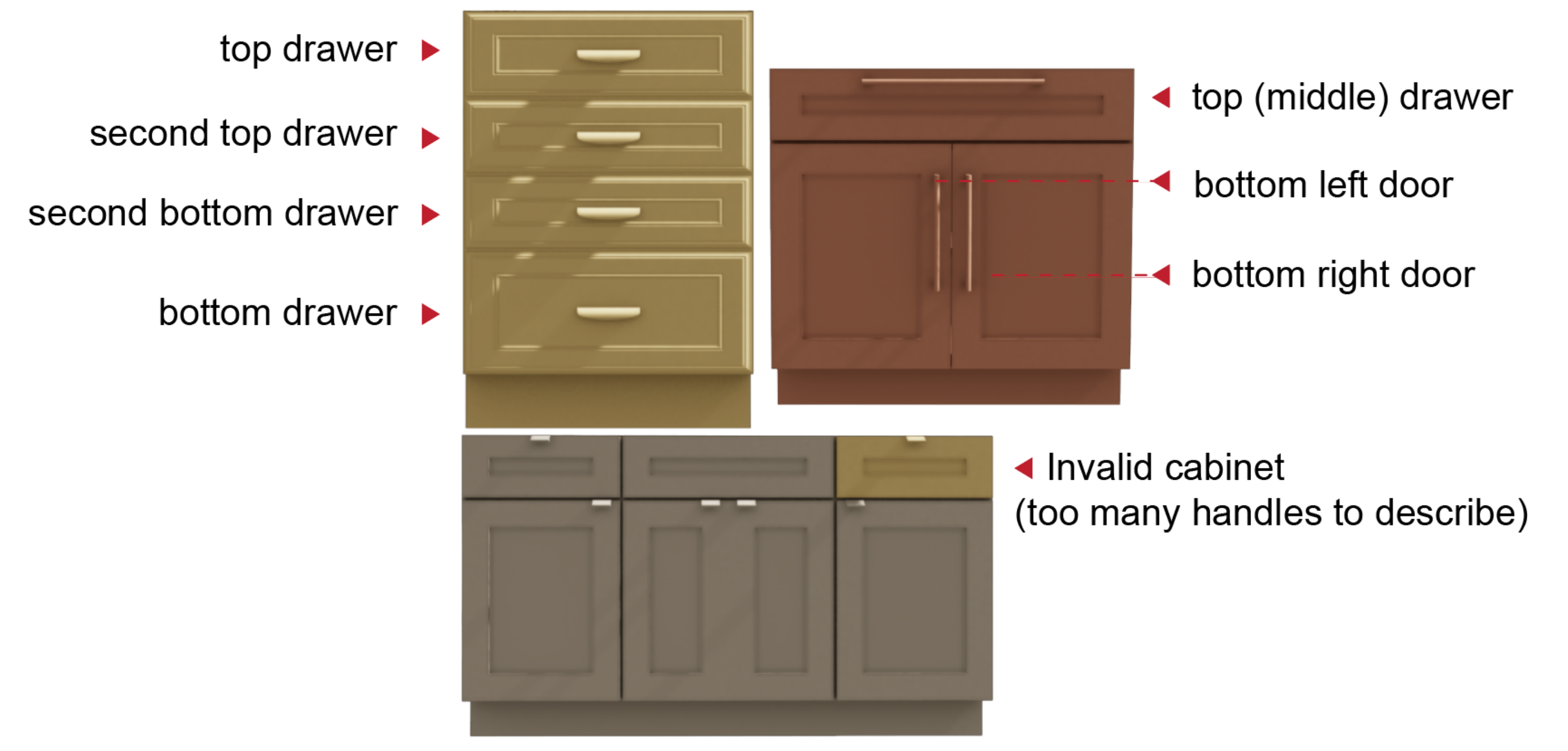}
    \caption{{\bf Language description examples.} On the top, language descriptions are validly generated. On the bottom, Algorithm~\ref{alg:lang} fails because of too many doors and drawers to describe.}
    \label{fig:lang_ex}
\end{figure}

We apply Algorithm 1 to generate template language as the spatial description for the drawers and doors on one cabinet. The key idea is to perform an iterative comparison between the target positions vertically and horizontally. However, this algorithm may fail to generate descriptions if the cabinet has too many doors and cabinets. For simplicity, we filter out these cabinets. Figure \ref{fig:lang_ex}
shows some generated instructions.

\begin{algorithm}[!ht]
\caption{Generate language instructions for one cabinet}\label{alg:lang}
\begin{algorithmic}
\Require Obtain the number of drawers and doors $n \geq 0$, and their positions $\{u_i~|~u_i = (x_i, y_i, z_i)\}_{i=1,2,...,n}$
\For{$i = 1,2,3,...,n$}
\If{$i \geq 2$}
    \State Compare $y_i$ with $\{y_1, ..., y_{i-1}\}$;
    \State Update description for $\{u_1,...,u_{i}\}$ with \{left, second left, middle, second right, and right\}
    \State Compare $z_i$ with $\{z_1, ..., z_{i-1}\}$;
    \State Update description for $\{u_1,...,u_{i}\}$ with \{top, second top, middle, second bottom, and bottom\};
\EndIf
\If{find the same description for two positions}
    \State \Return invalid cabinet.
\EndIf
\EndFor
\State \Return descriptions for $\{u_1,...,u_{n}\}$
\end{algorithmic}
\end{algorithm}

\textbf{Task Statistics.}
The original dataset contains $346$ pieces of the storage furniture. After filtering out low-quality meshes, URDF format~\cite{arrazate2017development} errors, and invalid cabinets from Algorithm~\ref{alg:lang}, we retain $174$ unique cabinets and a total number of $372$ doors and drawers with descriptions for training and testing. Table~\ref{tab:train_test_split} shows the train-test split.

\begin{table}[h]
\caption{Data split for training and testing}
\begin{center}
\begin{tabular}{l|c|cc}
\hline
               & Count & Train & Test \\ \hline
Cabinet drawer & 167   & 138 & 29               \\
Cabinet door   & 205   & 145 & 60             \\ \hline
Cabinet   & 174   & 135 & 39             \\ \hline

\end{tabular}
\end{center}
\label{tab:train_test_split}
\end{table}
\section{Model}

\begin{figure*}[t]
    \centering
    \includegraphics[width = 0.99\textwidth]{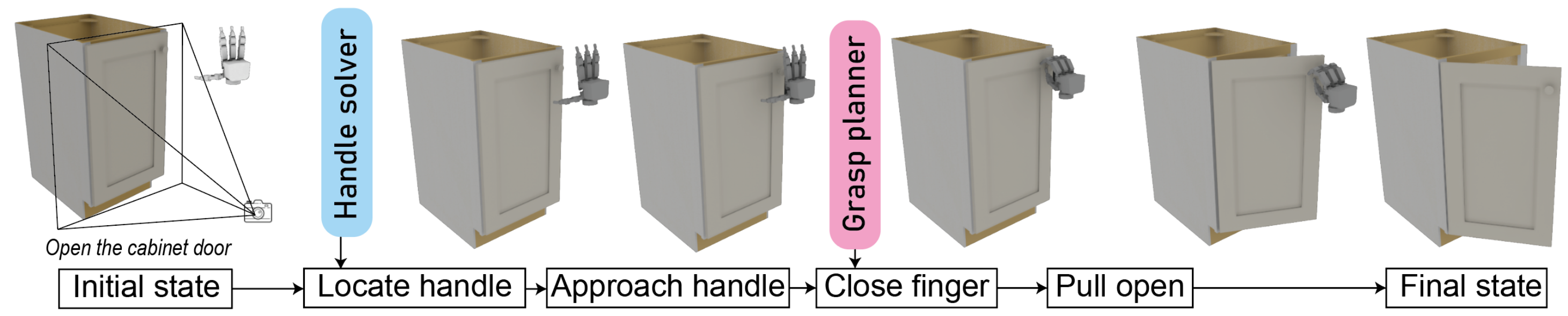}
    \caption{{\bf Multi-step planner to open a cabinet.} We apply such a planner to all hands. All the hands share the same handle solver to locate the handle position, while they differ in the grasp planner.}
    \label{fig:model}
\end{figure*}

Humans use visual perception to locate cabinets and understand verbal commands to determine which doors and drawers to open. After that, we open the target by controlling the movement of our hands and finger joints.

To address the challenges introduced by \textsc{OpenD}, we apply a multi-step planner to divide the overall task into several parts. As Figure \ref{fig:model} suggests, the \textbf{handle solver} copes with the challenge brought by image perception and language understanding, and the \textbf{grasp planner} precisely controls the movement of the finger joints. The reset parts of the planner are deterministic.

\subsection{The multi-step planner}

\textbf{Initial state.} The planner receives the inputs as RGB+D sensing $I$ of the cabinet and language $S$.

\textbf{Locate handle.} To accurately recognize the grasp position, a handle solver identifies and localizes the correct handle on the door or drawer from the image and language input. We characterize a handle by its bounding box $bbox = (y_0, y_1, z_1, z_1)$ and determine the its center $c$ and posture $r$:
\begin{align*}
    c &= ((y_0 + y_1) / 2, (z_0 + z_1) / 2) \\
    r &= vertical \text{ if } (z_1 - z_1) > (y_1 - y_0) \text{ else } horizontal
\end{align*}

\textbf{Approach handle.} Combining depth sensing on the $x$-axis and the bounding box, we recover the handle's world position relative to the camera. Then, we drive the hand to the front of the handle, preparing to grasp and pull.

\textbf{Close finger.} A grasp planner defines the movement of the finger joints $\{d_i\}$ to grasp to handle. 

\textbf{Pull open.} We drive  the robot hand  forward on the x-axis to pull open the target.

\textbf{Final state.} A task checker built in the \textsc{OpenD} determines the task's success based on the cabinet drawer's translation $\delta$ or cabinet door's rotation $\theta$. 

We define success conditions as: (1) the hand correctly opens the drawer or door described by the instruction; (2) the door open ratio $\theta / 180$ or the drawer open ratio $\delta / \text{(drawer length)}$ is greater than $20\%$.

\subsection{Handle solver}

 The handle solver helps locate the cabinet's correct handle to open, providing the RGB image $I$ and sentence $S$. We apply two compelling models that utilize deep neural networks to solve this problem.
 
\begin{figure}[h]
    \centering
    \includegraphics[width = 0.48\textwidth]{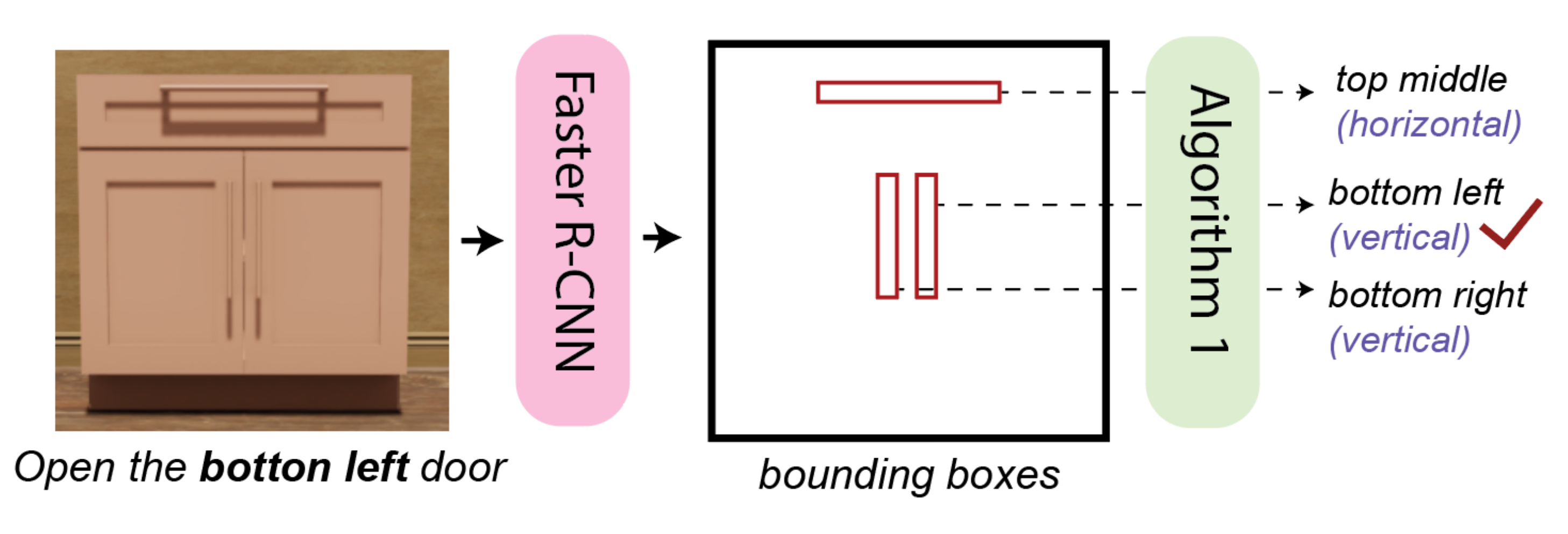}
    \caption{{\bf Training the handle solver with Faster-RCNN.} This model uses a two-step strategy to merge image and language information.}
    \label{fig:fasterrcnn}
\end{figure}

 The first model (see Figure \ref{fig:fasterrcnn}), \textbf{Faster R-CNN}~\cite{ren2015faster}, leverages the region proposal network to perform efficient object detection. From the image input, the Faster R-CNN model predicts the bounding boxes of all handles in the image. Then, we apply Algorithm \ref{alg:lang} to generate language instructions for predicted handles. Finally, we match the generated language instructions with sentence input $S$ to get the bounding box of the predicted handle.

\begin{figure}[h]
    \centering
    \includegraphics[width = 0.48\textwidth]{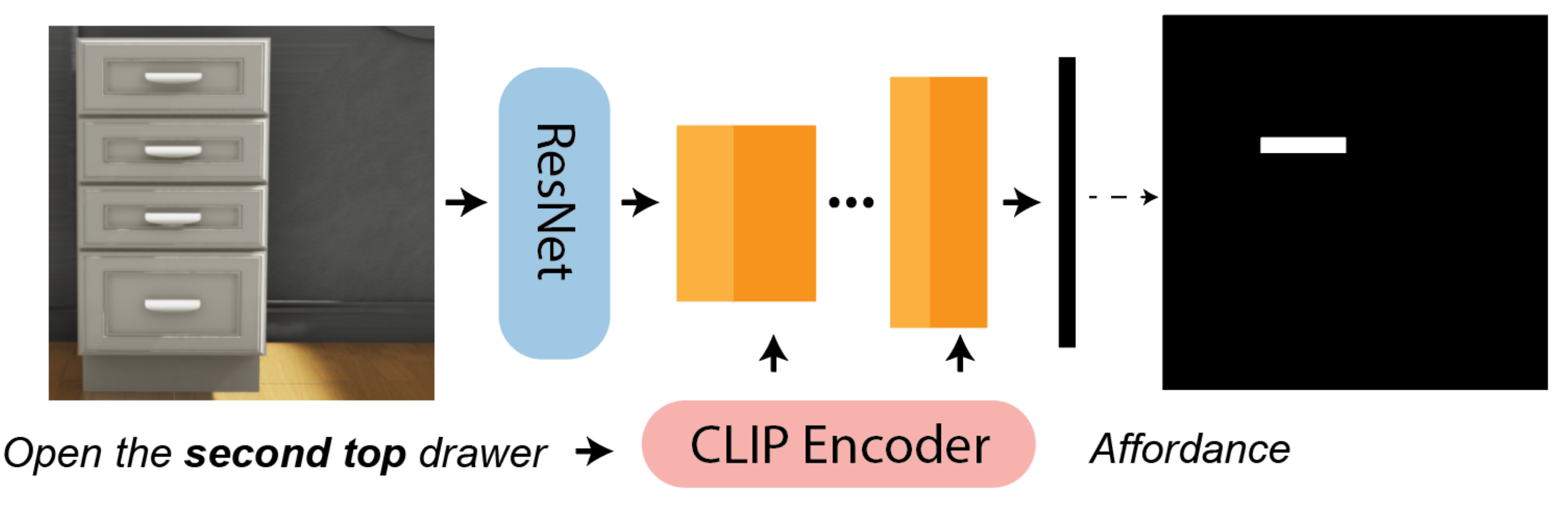}
    \caption{{\bf Training the handle solver with CLIPort.} This model gets the affordance of the bounding box directly from the image and language inputs}
    \label{fig:cliport}
\end{figure}

The second model (see Figure \ref{fig:cliport}), \textbf{CLIPort}~\cite{shridhar2022cliport},  combines learning generalizable semantic representations for vision and understanding necessary spatial information for fine-grained manipulation. CLIPort has a language-conditioned learning module to learn broad semantic knowledge (what) and the spatial precision (where) to transport. Since what to open (the handle) is known, we only apply half of the CLIPort to learn where to grasp according to the predicted affordance map.

\subsection{Grasp planner}
\begin{figure}[]
    \centering
    \includegraphics[width = 0.4\textwidth]{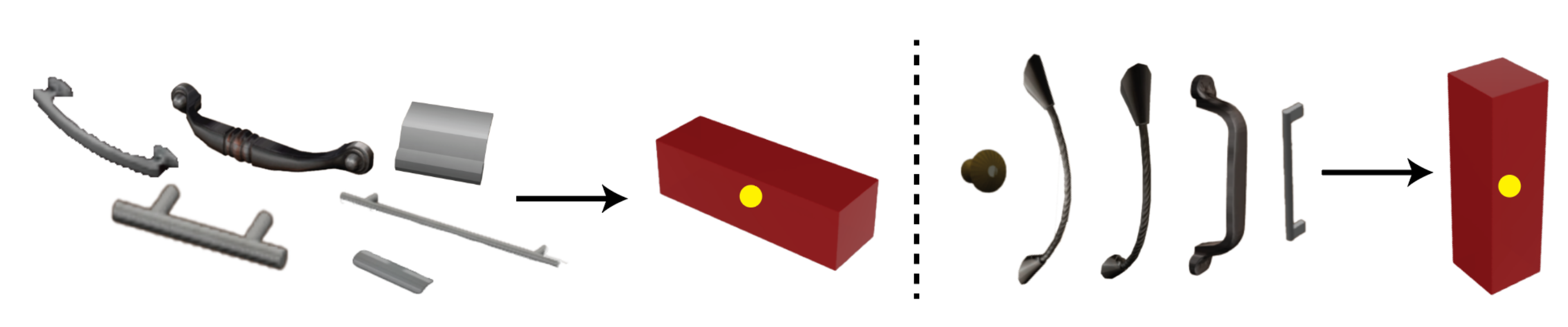}
    \caption{{\bf Handle simplification.} Our model ignores the handle's shape and only considers the center position and posture. Left: horizontal handles. Right: vertical handles. The simplification is only for the grasp searching algorithm and does not influence the simulation.}
    \label{fig:handle}
\end{figure}
After attending to a local region (bounding box) to decide where to grasp, a grasp planner drives the hand to close the fingers. We apply a context-independent grasp planner~\cite{saut2012efficient} to help determine the finger motion, by which we can determine where to place the
fingers on the handle. 

We simplify the original implementation in two ways. First, we only consider the object's center~$c$ and posture~$r$. Figure~\ref{fig:handle} shows some simplified examples of vertical and horizontal handles. Notice that this simplification is only for the grasp search algorithm; the detailed collision and friction are still fully enabled in simulation.  Second, we detach the hand from the robot arm and thus we do not consider the configuration that should be accessible to the robot.  


The grasp planner helps determine the final joint state of the hand. Thus, we calculate the intermediate joint state by interpolating the initial and final joint positions.


\section{Experiment}

We perform experiments in \omni~ to answer the following questions: 1) How to train the handler solver using the architecture of Faster R-CNN or CLIPort? 2) How does the performance of the grasp planner vary across different hands? 3) How generalizable is our multi-step planner when we put together the handler solver and the grasp planner? 

\textbf{Training the Faster R-CNN and CLIPort.}
To prepare training data, we first render cabinet images from the camera and retrieve bounding boxes for the handles from the engine. We introduce randomization to the training data to encourage more generalizable models. First, we randomly shift the camera position up to $10$ centimeters on each axis. Then, there is a $50\%$ chance to give a random color to the handle. The data augmentation procedure combining randomization results in $1740$ training images (resolution $256 \times 256$). The handler solver learns from the images and bounding boxes of handles projected to the image. We use $80\%$ of the images for training and $20\%$ for validation.

The Faster R-CNN model first loads a pre-trained backbone (ResNet50, MobileNetV3, or  Low-resolution MobileNet) on COCO (train2017) dataset~\cite{lin2014microsoft}. We fine tune the model for $20$ epochs on our training dataset based on regression loss (mean square error) of bounding box prediction. The Faster R-CNN models with the ResNet50, MobileNetV3, MobileNet backends reports training losses of $0.110$, $0.201$, and $0.207$, and validation losses of $0.158$,  $0.157$, and $0.153$. Even though the ResNet does not provide the best validation performance, we use it as the handle solver backend. We empirically find that the ResNet backend works well with Algorithm~\ref{alg:lang} to predict the bounding boxes, especially when we give a threshold of 0.8 to each bounding box detection score.



To train a CLIPort model, we first encode the language instructions using the CLIP text encoder~\cite{radford2021learning} and encode images using ResNet-18. Our predicting target is the affordance map highlighted by the bounding box. We fine tune the vision-language fusion layers of CLIPort on our training data via a binary cross-entropy loss.
We compare CLIPort models with different numbers of vision-language fusion layers. The models with four, three, and two layers perform similarly. After fine-tuning, they report the binary cross-entropy loss averaged on each pixel as $0.134$,  $0.135$, $0.132$ in training, and $0.146$, $0.148$, $0.152$ in validation. We use the 4-layer model as the second option for the handler solver.


\begin{figure}[h]
    \centering
    \includegraphics[width = 0.42\textwidth]{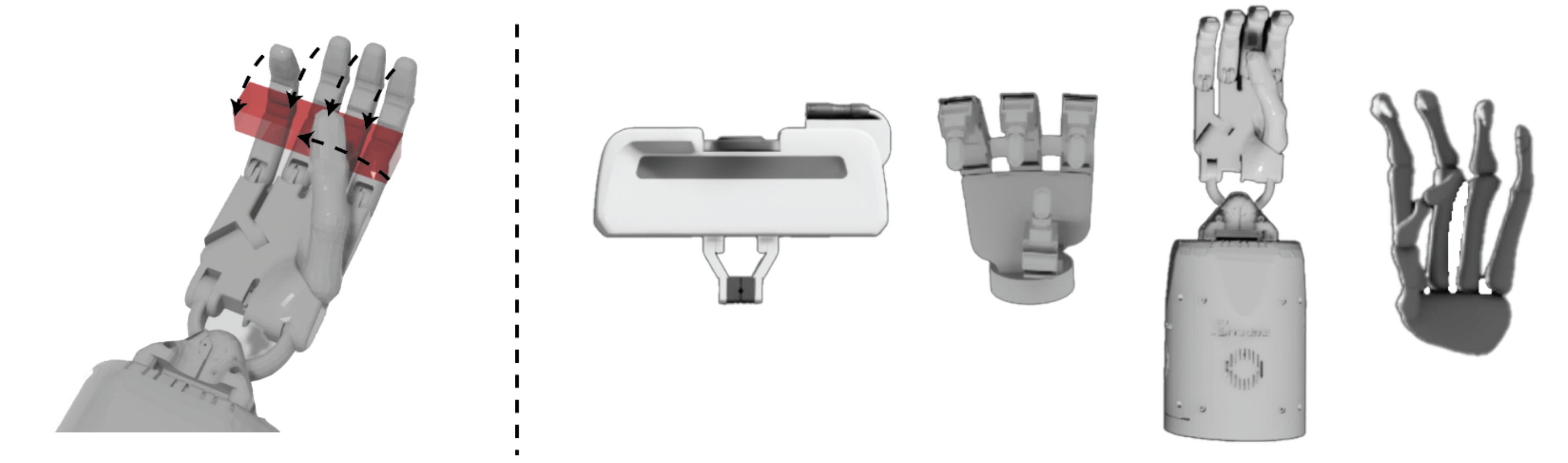}
    \caption{{\bf Grasp illustration.} Left: grasp planner illustration. We regard the handle as a cuboid and the grasp planner searches the  joint positions through curling fingers. Right: final grasping states deduced by the grasp planner.}
    \label{fig:hand_close}
\end{figure}

\textbf{Searching the grasp planner.}
The grasp planner defines the movement to close fingers for different types of hands. In our baseline, we design one grasp planner per hand. We take handles for the first three cabinets in the training dataset as samples for grasp searching. For Franka gripper, the grasp planner works simply by closing the two fingers. For the other three hands, the grasp planner searches the curling of each finger until the hand can successfully grasp the sample handles.

\begin{table}[h] 
\caption{Grasping success rate for different hands}
\begin{center}
\begin{tabular}{l|ccc}
\hline
               & Drawer     & Door    & Overall \\ \hline
Franka gripper & $91.6\%$ &   $58.5\%$      &  $73.4\%$ \\
Allegro hand   & $79.6\%$ &$66.2\%$&$72.8\%$\\
Shadow hand    &$92.3\%$ &$24.4\%$ & $54.8\%$ \\
Skeleton hand  & $50.3\%$ & $51.2\%$ & $50.8\%$  \\ \hline 
\end{tabular}
\end{center}
\label{tab:grasp}
\end{table}

\begin{table*}[t]
\caption{Task success rate for the multi-step planner.}
\begin{center}
\begin{tabular}{c|cccccc|cccccc}
\hline
\textbf{}      & \multicolumn{6}{c|}{\textbf{Faster R-CNN}}                                                                                          & \multicolumn{6}{c}{\textbf{CLIPort}}                                                                                                             \\ \cline{2-13} 
\textbf{}      & \multicolumn{2}{c|}{\textbf{Drawer}}                & \multicolumn{2}{c|}{\textbf{Door}}                  & \multicolumn{2}{c|}{\textbf{Overall}} & \multicolumn{2}{c|}{\textbf{Drawer}}                & \multicolumn{2}{c|}{\textbf{Door}}                  & \multicolumn{2}{c}{\textbf{Overall}} \\ \cline{2-13} 
\textbf{}      & \textbf{Train} & \multicolumn{1}{c|}{\textbf{Test}} & \textbf{Train} & \multicolumn{1}{c|}{\textbf{Test}} & \textbf{Train}     & \textbf{Test}    & \textbf{Train} & \multicolumn{1}{c|}{\textbf{Test}} & \textbf{Train} & \multicolumn{1}{c|}{\textbf{Test}} & \textbf{Train}    & \textbf{Test}    \\ \hline
Franka gripper &     $19.6\%$           & \multicolumn{1}{c|}{$20.7\%$}              &   $23.4\%$             & \multicolumn{1}{c|}{$13.3\%$}              &    $21.6\%$                &     $15.7\%$             &     $43.5\%$            & \multicolumn{1}{c|}{$35.9\%$ }              &    $19.3\%$             & \multicolumn{1}{c|}{$13.3\%$ }              &     $31.1\%$               &     $21.3\%$              \\
Allegro hand   & $36.2\%$       & \multicolumn{1}{c|}{$48.3\%$}      & $33.8\%$       & \multicolumn{1}{c|}{$21.7\%$}      & $35.0\%$           & $30.3\%$         & $59.7\%$       & \multicolumn{1}{c|}{$50.0\%$}      & $30.8\%$       & \multicolumn{1}{c|}{$20.0\%$}      & $45.0\%$          & $30.0\%$         \\
Shadow hand    & $47.1\%$       & \multicolumn{1}{c|}{$44.8\%$}      & $18.6\%$       & \multicolumn{1}{c|}{$8.3\%$}       & $24.7\%$           & $20.2\%$         & $17.3\%$       & \multicolumn{1}{c|}{$27.6\%$}      & $30.8\%$       & \multicolumn{1}{c|}{$14.8\%$}      & $17.6\%$          & $14.6\%$         \\
Skeleton hand  &  $15.9\%$   & \multicolumn{1}{c|}{$10.3\%$ }              &        $24.8\%$       & \multicolumn{1}{c|}{$16.7\%$}              &    $20.5\%$                &      $14.6\%$              &   $21.0\%$               & \multicolumn{1}{c|}{$17.2\%$  }              &      $26.9\%$            & \multicolumn{1}{c|}{$18.3\%$  }              &       $24.0\%$            &  $18.0\%$                \\ \hline
\end{tabular}
\label{tab:exp}
\end{center}
\end{table*}
\vspace{-2em}

Figure~\ref{fig:hand_close} depicts the illustration for grasp searching and the final grasp states. Table~\ref{tab:grasp} reports the corresponding grasping success rate for each hand type. 

If the hand is correctly placed at the grasping position, Franka gripper and Allegro hand can successfully open over $70\%$ of the targets, while the Shadow hand and Skeleton hand open around $50\%$ of them.  As the hand structure becomes more complex, the overall success rate drops, indicating that hands (such as the Shadow hand and Skeleton hand) require more precise control to handle the interaction with the handle. The experiment also shows that opening a cabinet door is more challenging than pulling open a drawer. Opening a door fails if the hand detaches unexpectedly from the handle during door rotation. 

\begin{figure}[h]
    \centering
    \includegraphics[width=.5\textwidth]{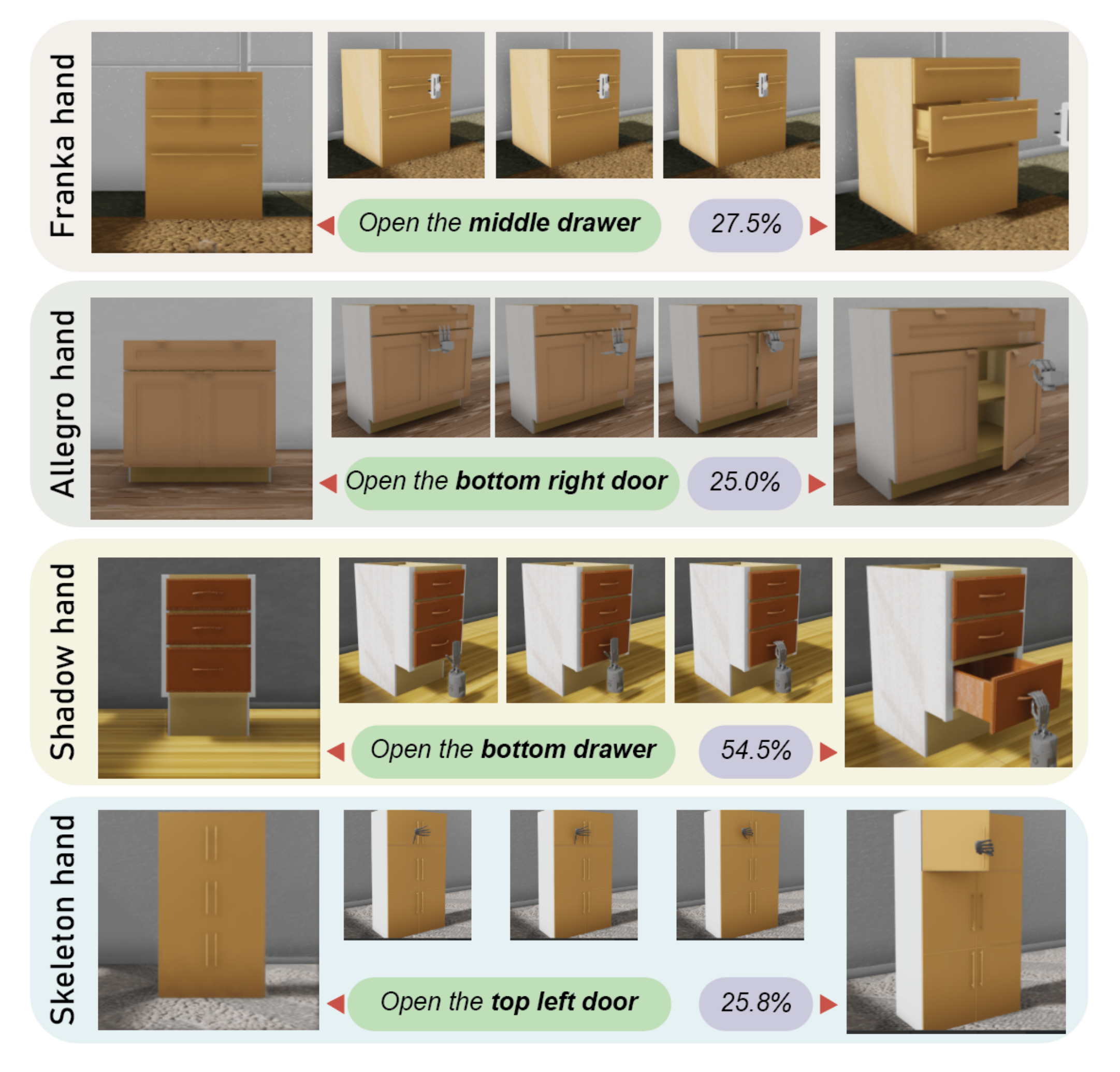}
    \caption{{\bf Task solving examples in the testing phase.} The green bubble highlights the language instruction, and the purple bubble shows the final open ratio. The leftmost image in each row, together with the instruction, is used as input for the handle solver.}
    \label{fig:test_example}
\end{figure}

\textbf{Overall results.}
Finally, we combine the well-trained handle solver and grasp planner to evaluate our baseline model on the test set. As shown in Table~\ref{tab:exp}, we obtain two sets of experimental results using two different handle solvers while keeping the same grasp planner for each hand. Figure~\ref{fig:test_example} plots one successful case for each hand during testing.

Comparing the experiment results, we find that the Allegro hand, retaining a hand shape while keeping a relatively simple structure, achieves the best testing performance (about $30\%$ success rate). The results also confirms that our model has a harder time opening doors than opening drawers. 

We also find that the Franka gripper needs more precise identification of the grasp position because of its smaller fingers. It can successfully grasp $73.4\%$ of the handles with ground-truth handle positions (see Table~\ref{tab:grasp}) but the performance drops drastically to $15.7\%$ (Faster R-CNN) and $21.3\%$ (CLIport) when recognition is involved.

From our model, the performance of the other two hands ($14.6\%$ to $20.2\%$ success rate) is not as good as that of the more straightforward Allegro hand ($30.0\%$). Only knowing the position and orientation of the hand seems insufficient for accurate control. 

\section{Discussion}

\textbf{Human performance.} We obtained a human evaluation of $10$ randomly sampled cabinet doors and cabinet drawer directives from the dataset. The experiment involved $5$ lab experts who randomly received initial Franka gripper positions and completed $100$ trajectories each for opening the drawer and the door. Each participant uses a gamepad controller. Before the experiment, the participants were allowed to familiarize themselves with the controller and the task. The participants obtained a high success rate of $92\%$ in opening the drawer and a $100\%$ success rate in opening the cabinet. The failure cases are mainly due to the loss of patience when grabbing the handle. This indicates that the tasks in \textsc{OpenD} are well-illustrated and shows that there is still plenty of room to improve the model.

\textbf{Robot configuration.} 
In a typical robot task, the placement of the robot predominantly determines whether the task can be accomplished. Especially for robots with fixed bases, the robot's initial position is critical. In \textsc{OpenD}, we first determine the rotation of the hand position. Then the hand can be treated as the end effector to derive the possible robot configuration. Given the state of the end effector, the whole robot status can be determined via inverse kinematics~\cite{narang2022factory}, sequential manipulation planning~\cite{stilman2010global}, or deep learning~\cite{tran2021identifying}.

\textbf{Task configuration.} Opening a cabinet becomes exceptionally complex as we randomize the set-up. In this study, we only cover a few essential random factors: object type, object location, and robot (hand) type, leaving a lot of unexplored random factors (e.g. interaction physics, camera properties).
For example, light intensity could substantially affect the experimental results. After weakening the light intensity by $30\%$, the affordance map obtained by the CLIPort model deviates significantly from the actual one, and the Faster R-CNN model has difficulty detecting bounding boxes. As a result, all hands perform poorly under the low-light condition.


\textbf{Other approaches.} To address the challenge, we also have tried two other approaches: transformer-based behavior cloning~\cite{mu2021maniskill} and offline reinforcement learning~\cite{levine2020offline}. However, the hundreds of demonstrations we collected for the human study are far from enough to train a transformer model. For \textsc{OpenD}, we set the physics update frequency to be $60$ Hz. Planning and behavior cloning last at least $5$ seconds on average to open the cabinet. Therefore, the algorithm for long-horizon planning and high-dimensional control for hands does not converge during reinforcement learning.

\section{Conclusion}

We introduce \textsc{OpenD}, a benchmark for learning to open  cabinet drawers and doors based on language instructions and visual inputs. \textsc{OpenD} sets up a challenging task via by simulating realistic physics and scenery. In particular, we avoid making a lot of simplifications to grasping configurations.

\textsc{OpenD} brings us closer to the community goal of language-driven robots that can perform accurate and stable interactions. The photo-realistic environment background and physics dynamics in \textsc{OpenD} are intended to narrow the gap between simulation and reality.



The proposed baseline approach demonstrates the difficulty of the task. We believe that better models can be constructed via exploiting hierarchical control, modular training, structured reasoning, and systematic planning. We are encouraged by the possibilities and challenges that \textsc{OpenD} introduces to the community.




\newpage
\bibliography{icra} 
 \bibliographystyle{ieeetr}

\end{document}